\pdfoutput=1

\documentclass[11pt]{article}

\usepackage{acl}

\usepackage{times}
\usepackage{latexsym}

\usepackage{CJKutf8}
\usepackage{multirow}
\usepackage{hhline}
\usepackage{booktabs}
\usepackage{graphicx}
\usepackage{enumitem}
\usepackage{tablefootnote}
\usepackage{amsmath}

\usepackage[T1]{fontenc}

\usepackage[utf8]{inputenc}

\usepackage{microtype}

\title{Parallel Data Helps Neural Entity Coreference Resolution}

\author{Gongbo Tang \\
  Beijing Language and Culture University \\
  \texttt{gongbo.tang@blcu.edu.cn} \\\And
  Christian Hardmeier \\
  IT University of Copenhagen  \\
  Uppsala University \\
  \texttt{chrha@itu.dk} \\}

\begin{document}
\maketitle
\begin{abstract}
Coreference resolution is the task of finding expressions that refer to the same entity in a text. Coreference models are generally trained on monolingual annotated data but annotating coreference is expensive and challenging. \citet{hardmeier-etal-2013-latent} have shown that parallel data contains latent anaphoric knowledge, but it has not been explored in end-to-end neural models yet. 
In this paper, we propose a simple yet effective model to exploit coreference knowledge from parallel data. In addition to the conventional modules learning coreference from annotations, we introduce an unsupervised module to capture cross-lingual coreference knowledge. 
Our proposed cross-lingual model achieves consistent improvements, up to 1.74 percentage points, on the OntoNotes 5.0 English dataset using 9 different synthetic parallel datasets. 
These experimental results confirm that parallel data can provide additional coreference knowledge which is beneficial to coreference resolution tasks. 
\end{abstract}

\section{Introduction}

Coreference resolution is the task of finding expressions, called mentions, that refer to the same entity in a text. Current neural coreference models are trained on monolingual annotated data, and their performance heavily relies on the amount of annotations \citep{lee-etal-2017-end,lee-etal-2018-higher,joshi-etal-2019-bert,joshi-etal-2020-spanbert}. Annotating such coreference information is challenging and expensive. Thus, annotation data is a bottleneck in neural coreference resolution.

\begin{figure}[t!]
\centering
        \includegraphics[totalheight=1.68cm]{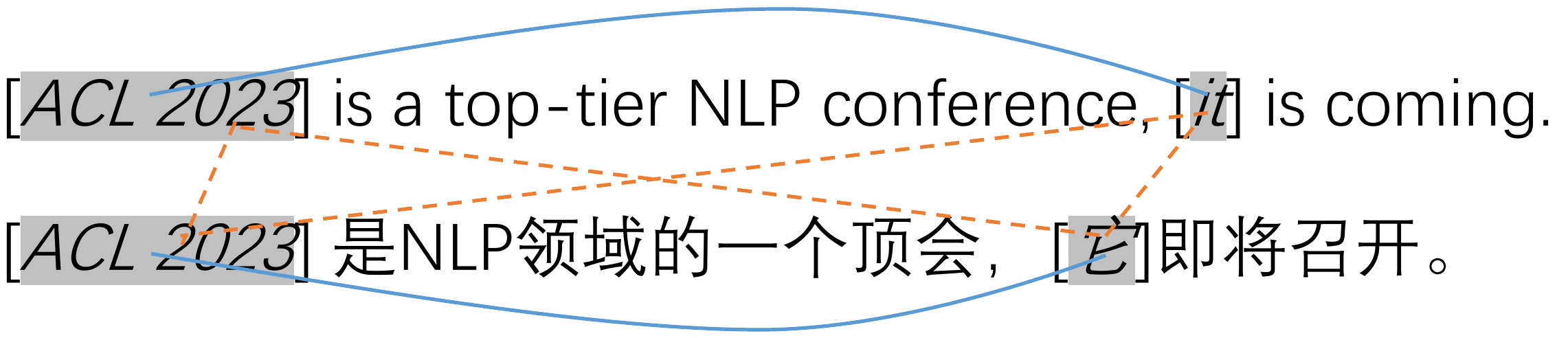}
    \caption{A coreference chain in an English--Chinese parallel sentence pair. Mentions in brackets are coreferential to each other. Links in blue are monolingual and dashed liks in orange are cross-lingual. }
    \label{fig:example}
\end{figure}  

\citet{hardmeier-etal-2013-latent} have explored parallel data in an unsupervised way and shown that parallel data has latent cross-lingual anaphoric knowledge. 
Figure~\ref{fig:example} shows a coreference chain in an English--Chinese parallel sentence pair. ``ACL 2023'', ``it'' in the English sentence, and ``ACL 2023'', \begin{CJK*}{UTF8}{gbsn}``它''\end{CJK*}(it) in the Chinese sentence are coreferential to each other. 
Compared to the two separate monolingual coreferential pairs: <ACL 2023, it>, <ACL 2023, \begin{CJK*}{UTF8}{gbsn}它\end{CJK*}>, there are four more cross-lingual coreferential pairs <it, ACL 2023>, <it, \begin{CJK*}{UTF8}{gbsn}它\end{CJK*}>, <ACL 2023, ACL 2023>, <ACL 2023, \begin{CJK*}{UTF8}{gbsn}它\end{CJK*}> in this parallel sentence pair. 
This cross-lingual coreference chain suggests that parallel multilingual data can provide extra coreferential knowledge compared to monolingual data which could be useful for training coreference models.

Parallel data has been applied to project coreference annotations in non-neural coreference models \citep{de2011can,rahman-ng-2012-translation,martins-2015-transferring,grishina-stede-2015-knowledge,novak-etal-2017-projection,grishina-stede-2017-multi}. Instead, we focus on neural coreference models and ask the following main research question: \textbf{Can parallel data advance the performance of coreference resolution on English, where a relatively large amount of annotations are available? }


We propose a cross-lingual model which exploits cross-lingual coreference knowledge from parallel data. 
Our model is based on the most popular neural coreference model (\citealp{lee-etal-2018-higher}), which consists of an encoder, a mention span scorer, and a coreference scorer. We extend these three modules, which are applied to the source-side data, with a target-side encoder and adapters for the mention span scorer and the coreference scorer, allowing these to resolve cross-lingual coreference. 
As there is no annotated cross-lingual coreference data, the model computes the coreference scores between target spans and source spans without any supervision. 
We conduct experiments on the most popular OntoNotes 5.0 English dataset \citep{pradhan-etal-2012-conll}. 
Given the English data, we generate 9 different synthetic parallel datasets with the help of pretrained neural machine translation (NMT) models. The target languages consist of Arabic, Catalan, Chinese, Dutch, French, German, Italian, Russian, and Spanish. 
The experimental results show that our cross-lingual models achieve consistent improvements, which confirms that parallel data helps neural entity coreference resolution.

\section{Related Work}

\citet{lee-etal-2017-end} first propose end-to-end neural coreference models (\textit{neural-coref}) and achieve better performance on the OntoNotes English dataset compared to previous models. Most current neural coreference models are based on \textit{neural-coref} and replace the statistic word embeddings used by \citet{lee-etal-2017-end} with contextualized word embeddings from ELMo \citep{peters-etal-2018-deep}, BERT \citep{devlin-etal-2019-bert}, SpanBERT \citep{joshi-etal-2020-spanbert}, etc. \citep{lee-etal-2018-higher,joshi-etal-2019-bert,joshi-etal-2020-spanbert}. 

\textit{Neural-coref} only models the relation between pairs of mentions. Many studies propose to consider entity-level information while predicting clusters \citep{lee-etal-2018-higher,kantor-globerson-2019-coreference,xu-choi-2020-revealing}. However, \citet{xu-choi-2020-revealing} find that these models considering higher-order inference are not significantly better or even worse. Instead, the observed differences can be explained by the powerful performance of SpanBERT.  

Because these models are expensive in terms of memory and time, especially when using higher-dimensional representations. \citet{xia-etal-2020-incremental} and \citet{toshniwal-etal-2020-learning} propose models that only keep a limited number of entities in the memory, without much performance drop. 
\citet{kirstain-etal-2021-coreference} introduce a \textit{start-to-end} model where the model computes mention and antecedent scores only through bilinear functions of span boundary representations. 
To cope with the enormous number of spans, \citet{dobrovolskii-2021-word} proposes a word-level coreference model, where the model first considers the coreference links between single words, and then reconstruct the word spans. 

All these models are trained on monolingual coreference annotations. In this paper, we introduce a simple model building on the top of \textit{neural-coref}, which exploits cross-lingual coreference from  parallel data in an unsupervised way.

\begin{figure}[htbp]
\centering
        \includegraphics[totalheight=4.5cm]{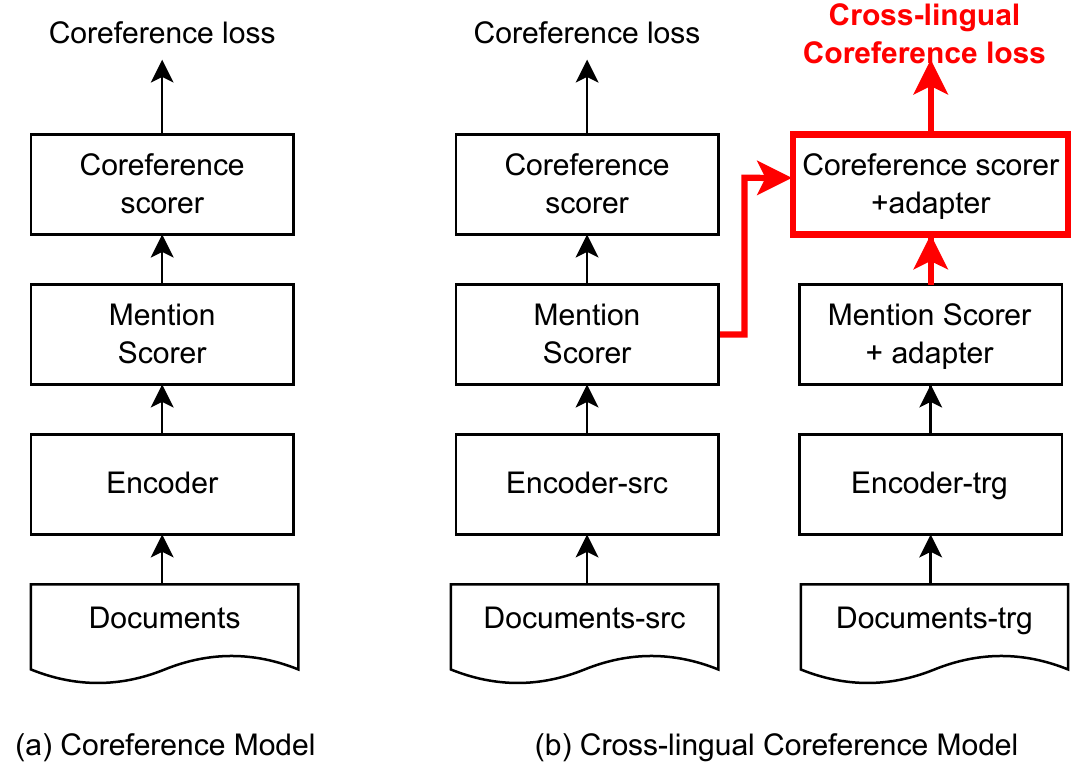}
    \caption{Overview of (a) the conventional monolingual coreference model  and (b) our cross-lingual coreference model using synthetic parallel data. The main differences are marked in red. The red block is a cross-lingual coreference scorer which is expected to capture cross-lingual coreference knowledge.}
    \label{fig:model}
\end{figure}

\section{Coreference Models}

\subsection{\textit{neural-coref}}
Most neural coreference models are variants of \textit{neural-coref} \citep{lee-etal-2017-end}, whose structure is illustrated in Figure~\ref{fig:model} (a). It consists of a text encoder, a mention scorer, and a coreference scorer. The final coreference clusters are predicted based on the scores of these modules. 

Given a document, the encoder first generates representations for each token. Then the model creates a list of spans, varying the span width.\footnote{The number of generated spans is decided by hyper-parameters, i.e., the maximum width of a span, the ratio of entire span space, the maximum number of spans.} Each span representation is the concatenation of 1) the first token representation, 2) the last token representation, 3) the span head representation, and 4) the feature vector, where the span head representation is learned by an attention mechanism \citep{bahdanau15joint} and the feature vector encodes the size of the span. 
Then the mention scorer, a feed-forward neural network, assigns a score to each span. 
Afterwards, the coreference scorer computes how likely it is that a mention refers to each of the preceding mentions. 

During training, given a span $i$, the model predicts a set of possible antecedents $\mathcal{Y}=\{\epsilon, 1, \dots, i - 1\}$, a dummy antecedent $\epsilon$ and preceding spans. The model generates a probability distribution $P(y_{i})$ over antecedents for the span $i$, as shown in Equation~\ref{eq:coref} below. $s(i,j)$ denotes the coreference score between span pair $i$ and $j$. 
The coreference loss is the marginal log-likelihood of the correct antecedents. 
During inference, the model first recognizes potential antecedents for each mention, then it predicts the final coreference clusters. More specifically, given a mention, the model considers the preceding mention with the highest coreference score as the antecedent.  

\begin{equation} \label{eq:coref}
P(y_{i}) = \frac{e^{s(i,y_{i})}}{\sum_{y' \in \mathcal{Y}(i)}e^{s(i,y')}} 
\end{equation}

\subsection{Cross-Lingual Model}

We hypothesize that parallel data can provide additional coreference information which benefits learning coreference. 
As there is no supervision to the target-side and cross-lingual modelling, we attempt to transfer the source-side learned parameters to the target-side unsupervised modules by adding additional adapters, which has been shown efficient and effective \cite{pmlr-v97-houlsby19a}. 
Therefore, we extend \textit{neural-coref} by introducing a target-side encoder, adapters for target-side mention scorer, and cross-lingual coreference scorer, where each adapter is a one-layer feed-forward neural network with 500 hidden nodes. 
The overview of our cross-lingual model is shown in Figure~\ref{fig:model} (b). 

For the target-side, we can use a shared cross-lingual encoder or a target-side monolingual encoder. 
The coreference scorer computes coreference scores between target-side spans and source-side spans. This is the key component to learn cross-lingual coreference knowledge. 
The strategy we follow is the same as that in \textit{neural-coref} during inference: Given a source mention, the target mention with the highest coreference score is considered as the corresponding cross-lingual antecedent. This component serves to capture latent coreference information. 
During training, as source-side modules are shared across languages, source-side parameters are jointly updated when optimizing the cross-lingual coreference loss. 

There is no specific range for antecedents in the cross-lingual setting. Thus, we introduce a restriction to target-side antecedents, where the cross-lingual antecedent's position number in the target sentence should not surpass the source mention's position number in the source sentence more than 50. This pruning can make the model more efficient and effective. 

Say the model has predicted a source mention list $M_{s}$: $\{m_{s_{1}}, m_{s_{2}},\dots, m_{s_{m}}\}$ and a target mention list $M_{t}$: $\{m_{t_{1}}, m_{t_{2}},\dots, m_{t_{n}}\}$.  
The model has also generated a two-dimensional coreference score matrix, where $s_{ij}$ represents the coreference score between $m_{s_{i}}$ and $m_{t_{j}}$. 
We denote $\mathcal{Y}(i)$ as the possible antecedent set of the source mention $i$. 
The cross-lingual coreference loss is defined in Equation~\ref{eq:agreement}, 
where $\hat{j}= \underset{j \in \mathcal{Y}(i)}{\arg \max}\ s_{ij}$ for a given $i$.\footnote{We assume that there should be at least one antecedent on the other side for each mention, either the translation of the mention or a translation of its antecedent. In practice, the quality of synthetic parallel data is not guaranteed which introduces noise. On the other hand, synthetic data may actually be more parallel than natural translations. }
\begin{equation} \label{eq:agreement}
\mathcal{L}_{x} = \textstyle \sum_{i=1}^{m}e^{-s_{i\hat{j}}} 
\end{equation}

During training, the model learns to minimize both the coreference loss and the cross-lingual coreference loss $\mathcal{L}_{x}$ with a ratio $1:1$. 
During inference, we only employ the source-side modules, which are trained with coreference supervision and latent cross-lingual coreference knowledge, to predict coreference clusters.

\begin{table*}[ht]
\centering
\scalebox{0.69}{
\begin{tabular}{lcccccccccccccccccc}
\toprule
\multirow{2}{*}{Data} &\multirow{2}{*}{$F1_{mention}$} & & \multicolumn{3}{c}{$MUC$} & &\multicolumn{3}{c}{$B^{3}$} & &\multicolumn{3}{c}{$CEAF_{e}$} & &\multirow{2}{*}{$F1_{avg}$} & & \multirow{2}{*}{$\Delta$ F1} \\
& & &R & P & F1 & &R & P & F1 & &R & P & F1 && & \\
\midrule
English&85.42&&80.31&81.40&80.85&&71.31&70.92&71.10&&65.81&70.97&68.30&&73.42&&\phantom{1.1}0\\
\midrule
English--Arabic&86.13&&81.73&81.80&81.77&&72.91&71.77&72.34&&67.85&71.53&69.64&&74.58&&1.16\\
English--Catalan&86.17&&81.38&82.36&81.87&&72.55&72.75&72.65&&67.77&72.19&69.91&&74.81&&1.39\\
English--Chinese&86.02&&81.16&82.43&81.78&&71.91&72.74&72.32&&66.96&72.17&69.47&&74.53&&1.11\\
English--Dutch&\textbf{86.29}&&81.53&\textbf{82.84}&\textbf{82.18}&&72.67&\textbf{73.31}&\textbf{72.99}&&\textbf{68.36}&\textbf{72.41}&\textbf{70.33}&&\textbf{75.16}&&\textbf{1.74}\\
English--French&85.93&&81.12&82.15&81.63&&72.06&72.36&72.20&&67.36&71.31&69.28&&74.37&&0.95\\
English--German&86.02&&81.86&81.28&81.56&&73.06&70.82&71.92&&67.42&70.93&69.14&&74.20&&0.78\\
English--Italian&86.13&&81.71&82.09&81.90&&72.82&72.09&72.45&&67.73&71.60&69.61&&74.65&&1.23\\
English--Russian&86.17&&\textbf{82.38}&81.31&81.84&&\textbf{73.75}&70.62&72.15&&67.94&71.12&69.49&&74.50&&1.08\\
English--Spanish&86.21&&81.72&81.88&81.80&&72.62&71.88&72.25&&67.88&71.11&69.45&&74.50&&1.08\\
\bottomrule
\end{tabular}
}
\caption{\label{table-result-en} F1 scores on mention detection ($F1_{mention}$) and coreference resolution ($F1_{avg}$) of the monolingual model trained on English and cross-lingual models trained on 9 different synthetic parallel datasets. $\Delta$ F1 is the improvement over the monolingual model. Bold numbers are the best scores in each column. $F1_{avg}$ scores of all the cross-lingual models are statistically significant (t-test, $p<0.05$). }
\end{table*}

\section{Experiments}

Due to the page limit, we leave our experimental settings in Appendix~\ref{appdex:setting}. 

\subsection{Data}

We experiment with the OntoNotes 5.0 
English dataset. The number of documents for training, development, and test is 2,802, 343, and 348, respectively. The data is originally from newswire, magazines, broadcast news, broadcast conversations, web, conversational speech, and the Bible. It has been the benchmark dataset for coreference resolution since it is released. The annotation in OntoNotes covers both entities and events, but with a very restricted definition of events. Noun phrases, pronouns, and head of verb phrases are considered as potential mentions. 
Singleton clusters\footnote{An entity cluster that only contains a single mention.} are not annotated in OntoNotes. 

Given the English data, we use open access pretrained NMT models released by Facebook and the Helsinki NLP group to generate synthetic parallel data \citep{wu2019pay,ng-etal-2019-facebook,tiedemann-thottingal-2020-opus}. 

The input to monolingual models are the English data and the inputs to cross-lingual models are these parallel data. They have the same amount of data entries. 
These parallel data have the same coreference annotations as the data fed into monolingual models, the only difference is that the English data is paired with its target translations, and there are no annotations in the translations at all.

\subsection{Experimental Results}

Table~\ref{table-result-en} shows the detailed scores of each model on the OntoNotes 5.0 English test set. 
Compared to the baseline model, which is trained only on English data, our cross-lingual model trained on different synthetic parallel datasets achieves consistent and statistically significant (t-test, $p<0.05$) improvements, varying from 0.78 to 1.74 percentage points. 
The model trained on English--Dutch achieves the best F1 performance on coreference resolution. The model trained on English--Russian achieves the best recall score on $MUC$ and $B^{3}$. 

It is interesting to see that the model trained on English--German achieves the least improvement, although German together with Dutch are closer to English compared to other languages. Meanwhile, the models trained on English--Arabic, English--Chinese, English--Russian obtain moderate improvements, even though Arabic, Chinese, and Russian are more different from English. Given the reported BLEU scores of the pre-trained NMT models, we find that the improvements do not correlate with the quality of generated translations. 

In addition to the results on coreference resolution, we also report the mention detection results, which are based on mention scores, i.e., the outputs of mention scorers. Models trained on parallel data are consistently superior to the monolingual model, and the model trained on English--Dutch gets the best F1 score of 86.29. 
We can tell that models with a higher mention detection F1 score do not always achieve higher coreference F1 score. There is no consistency across different language pairs, so the improvements are not merely from better mention detection performance, namely, memorizing mentions. 

As Table~\ref{table-result-en} shows, our cross-lingual model, which exploits parallel data, is superior to the model trained only on monolingual data. This confirms that parallel data can provide additional coreference knowledge to coreference models, which is beneficial to coreference modelling, even if the parallel data is synthetic and noisy.\footnote{We also conduct preliminary experiments with parallel data from multiple language pairs, concatenating the parallel data of EN-DE, EN-ES, EN-IT, EN-NL, and EN-RU five language pairs. Our proposed cross-lingual model achieves better performance compared to using data from one single language pair, showing the capability of our model to work with multiple parallel data. 

}

\section{Analysis}

\subsection{Unsupervised Cross-Lingual Coreference}

To explore whether the unsupervised module can capture cross-lingual coreference information, we check the cross-lingual mention pairs predicted by the cross-lingual coreference scorer. 

\textit{ParCorFull} \citep{lapshinova-koltunski-etal-2018-parcorfull} is an English--German parallel corpus annotated with coreference chains. We first feed the data to the model and let the model predict English--German mention pairs. We go through the these pairs quickly and find that some of these pairs are coreferential, some of these pairs are translation pairs, but most of them are irrelevant. As the coreference chains in English and German are not aligned, we cannot conduct quantitative evaluation. 

Alternatively, we evaluate the ability of the model to capture cross-lingual coreference knowledge using a synthetic mention pair set: an English--English mention pair set. Now we have ``aligned'' coreference chains, and we can evaluate the mention pairs automatically. 
Specifically, we first train a cross-lingual model with English--English synthetic data, and we then feed the OntoNotes English validation set to the model, both the source and target sides, to predict English--English mention pairs. 

The model predicts 18,154 pairs in total, including 131 mention pairs that are the same mention, 1,257 mention pairs that are coreferential, and 758 mention pairs with the same surface. This indicates that the model is able to resolve some cross-lingual coreference. However, since the cross-lingual module is trained without any supervision, most of predicted mention pairs are not coreferential. 

Table~\ref{x-lingual-mention-pair} shows some correctly predicted coreferential mention pairs, in English--English and English--German settings. We can tell that our cross-lingual models are not simply generating a pair of two identical mentions, but coreferential mentions as well, which is different from word alignment. These mention pairs support our hypothesis that the cross-lingual model can capture cross-lingual coreference knowledge.

\begin{table}[ht]
\centering
\scalebox{0.62}{
\begin{tabular}{ll}
\toprule
Source Mentions(English) & Target Mentions(English/German)\\
\midrule
Hong Kong&the city 's\\
It & the Supreme Court \\
he&28-jähriger Koch (28-Year-Old Chef)\\
The 19-year-old American gymnast &Simone Biles\\
\bottomrule
\end{tabular}
}
\caption{\label{x-lingual-mention-pair} Examples of correct coreferential mention pairs predicted by the cross-lingual coreference model, in English--English, English--German settings. }
\end{table}

\subsection{Separate Monolingual Encoders}
\label{ssec:sep-encoders}

Multilingual pretrained models suffer from the curse of multilinguality which makes them less competitive as monolingual models. Thus, we test the robustness of our model with separate encoders, i.e., we replace the unified cross-lingual encoder (XLM-R) with two separate monolingual encoders. The baseline is a monolingual model trained with SpanBERT, and the cross-lingual model is trained with SpanBERT and BERT on source- and target-side text, on the English--German synthetic dataset. 

Our experimental results show that models employing SpanBERT perform much better, which is consistent with previous findings by \citet{joshi-etal-2020-spanbert}. 
The monolingual model achieves 77.26 F1 score on the OntoNotes 5.0 English test set. Our cross-lingual model obtains an even higher F1 score, 77.79, which is statistically significant (t-test, p=0.044). 
Thus, our proposed model is applicable to settings with separate monolingual encoders. 

The improvement on SpanBERT is smaller than that on XLM-R. One explanation is that SpanBERT is already very powerful and parallel data provides less additional knowledge. Another explanation is that the target-side encoder, a BERT model, is much weaker than SpanBERT, which makes it harder to learn the cross-lingual coreference.

\section{Conclusions and Future Work}

In this paper, we introduce a simple yet effective cross-lingual coreference resolution model to learn coreference from synthetic parallel data. Compared to models trained on monolingual data, our cross-lingual model achieves consistent improvements, varying from 0.78 to 1.74 percentage points, on the OntoNotes 5.0 English dataset, which confirms that parallel data benefits neural coreference resolution. 

We have shown that the unsupervised cross-lingual coreference module can learn limited coreference knowledge. In future work, it would be interesting if we can provide the model some aligned cross-lingual coreference knowledge for supervision, to leverage parallel data better. 

\section*{Limitations}
We expect that our cross-lingual models have learnt some coreference knowledge on the target languages and we conduct experiments on some languages in zero-shot settings.  However, we do not get consistent and significant improvements compared to monolingual models. This should be further investigated which potentially helps languages with few or no coreference annotations. 
Compared to monolingual models, our cross-lingual model improves the source-side coreference resolution but it requires almost two times GPU memory during training. Thus, this model architecture imposes restrictions on using larger pretrained models given limited resources.

\section*{Acknowledgments}
We thank all reviewers for their valuable and insightful comments. This project is mainly funded by the Swedish Research Council (grant 2017-930), under the project Neural Pronoun Models for Machine Translation. 
GT is also supported by Science Foundation of Beijing Language and Culture University (supported by “the Fundamental Research Funds for the Central Universities”) (22YBB36). 
We also acknowledge the CSC – IT Center for Science Ltd., for computational resources, with the help of J\"org Tiedemann.

\bibliography{anthology,custom}

\begin{thebibliography}{30}
\expandafter\ifx\csname natexlab\endcsname\relax\def\natexlab#1{#1}\fi

\bibitem[{Bagga and Baldwin(1998)}]{bagga1998algorithms}
Amit Bagga and Breck Baldwin. 1998.
\newblock Algorithms for scoring coreference chains.
\newblock In \emph{The first international conference on language resources and
  evaluation workshop on linguistics coreference}, volume~1, pages 563--566.

\bibitem[{Bahdanau et~al.(2015)Bahdanau, Cho, and Bengio}]{bahdanau15joint}
Dzmitry Bahdanau, Kyunghyun Cho, and Yoshua Bengio. 2015.
\newblock \href {https://arxiv.org/pdf/1409.0473.pdf} {Neural machine
  translation by jointly learning to align and translate}.
\newblock In \emph{Proceedings of the 3rd International Conference on Learning
  Representations}, San Diego, USA.

\bibitem[{de~Souza and Or{\u{a}}san(2011)}]{de2011can}
Jos{\'e} Guilherme~Camargo de~Souza and Constantin Or{\u{a}}san. 2011.
\newblock \href {https://doi.org/https://doi.org/10.1007/978-3-642-25917-3_6}
  {Can projected chains in parallel corpora help coreference resolution?}
\newblock In \emph{Anaphora Processing and Applications}, pages 59--69, Berlin,
  Heidelberg. Springer.

\bibitem[{Devlin et~al.(2019)Devlin, Chang, Lee, and
  Toutanova}]{devlin-etal-2019-bert}
Jacob Devlin, Ming-Wei Chang, Kenton Lee, and Kristina Toutanova. 2019.
\newblock \href {https://doi.org/10.18653/v1/N19-1423} {{BERT}: Pre-training of
  deep bidirectional transformers for language understanding}.
\newblock In \emph{Proceedings of the 2019 Conference of the North {A}merican
  Chapter of the Association for Computational Linguistics: Human Language
  Technologies, Volume 1 (Long and Short Papers)}, pages 4171--4186,
  Minneapolis, Minnesota. Association for Computational Linguistics.

\bibitem[{Dobrovolskii(2021)}]{dobrovolskii-2021-word}
Vladimir Dobrovolskii. 2021.
\newblock \href {https://aclanthology.org/2021.emnlp-main.605} {Word-level
  coreference resolution}.
\newblock In \emph{Proceedings of the 2021 Conference on Empirical Methods in
  Natural Language Processing}, pages 7670--7675, Online and Punta Cana,
  Dominican Republic. Association for Computational Linguistics.

\bibitem[{Grishina and Stede(2015)}]{grishina-stede-2015-knowledge}
Yulia Grishina and Manfred Stede. 2015.
\newblock \href {https://doi.org/10.18653/v1/W15-3403} {Knowledge-lean
  projection of coreference chains across languages}.
\newblock In \emph{Proceedings of the Eighth Workshop on Building and Using
  Comparable Corpora}, pages 14--22, Beijing, China. Association for
  Computational Linguistics.

\bibitem[{Grishina and Stede(2017)}]{grishina-stede-2017-multi}
Yulia Grishina and Manfred Stede. 2017.
\newblock \href {https://doi.org/10.18653/v1/W17-1506} {Multi-source annotation
  projection of coreference chains: assessing strategies and testing
  opportunities}.
\newblock In \emph{Proceedings of the 2nd Workshop on Coreference Resolution
  Beyond {O}nto{N}otes ({CORBON} 2017)}, pages 41--50, Valencia, Spain.
  Association for Computational Linguistics.

\bibitem[{Hardmeier et~al.(2013)Hardmeier, Tiedemann, and
  Nivre}]{hardmeier-etal-2013-latent}
Christian Hardmeier, J{\"o}rg Tiedemann, and Joakim Nivre. 2013.
\newblock \href {https://aclanthology.org/D13-1037} {Latent anaphora resolution
  for cross-lingual pronoun prediction}.
\newblock In \emph{Proceedings of the 2013 Conference on Empirical Methods in
  Natural Language Processing}, pages 380--391, Seattle, Washington, USA.
  Association for Computational Linguistics.

\bibitem[{Houlsby et~al.(2019)Houlsby, Giurgiu, Jastrzebski, Morrone,
  De~Laroussilhe, Gesmundo, Attariyan, and Gelly}]{pmlr-v97-houlsby19a}
Neil Houlsby, Andrei Giurgiu, Stanislaw Jastrzebski, Bruna Morrone, Quentin
  De~Laroussilhe, Andrea Gesmundo, Mona Attariyan, and Sylvain Gelly. 2019.
\newblock \href {https://proceedings.mlr.press/v97/houlsby19a.html}
  {Parameter-efficient transfer learning for {NLP}}.
\newblock In \emph{Proceedings of the 36th International Conference on Machine
  Learning}, volume~97 of \emph{Proceedings of Machine Learning Research},
  pages 2790--2799. PMLR.

\bibitem[{Joshi et~al.(2020)Joshi, Chen, Liu, Weld, Zettlemoyer, and
  Levy}]{joshi-etal-2020-spanbert}
Mandar Joshi, Danqi Chen, Yinhan Liu, Daniel~S. Weld, Luke Zettlemoyer, and
  Omer Levy. 2020.
\newblock \href {https://doi.org/10.1162/tacl_a_00300} {{S}pan{BERT}: Improving
  pre-training by representing and predicting spans}.
\newblock \emph{Transactions of the Association for Computational Linguistics},
  8:64--77.

\bibitem[{Joshi et~al.(2019)Joshi, Levy, Zettlemoyer, and
  Weld}]{joshi-etal-2019-bert}
Mandar Joshi, Omer Levy, Luke Zettlemoyer, and Daniel Weld. 2019.
\newblock \href {https://doi.org/10.18653/v1/D19-1588} {{BERT} for coreference
  resolution: Baselines and analysis}.
\newblock In \emph{Proceedings of the 2019 Conference on Empirical Methods in
  Natural Language Processing and the 9th International Joint Conference on
  Natural Language Processing (EMNLP-IJCNLP)}, pages 5803--5808, Hong Kong,
  China. Association for Computational Linguistics.

\bibitem[{Kantor and Globerson(2019)}]{kantor-globerson-2019-coreference}
Ben Kantor and Amir Globerson. 2019.
\newblock \href {https://doi.org/10.18653/v1/P19-1066} {Coreference resolution
  with entity equalization}.
\newblock In \emph{Proceedings of the 57th Annual Meeting of the Association
  for Computational Linguistics}, pages 673--677, Florence, Italy. Association
  for Computational Linguistics.

\bibitem[{Kirstain et~al.(2021)Kirstain, Ram, and
  Levy}]{kirstain-etal-2021-coreference}
Yuval Kirstain, Ori Ram, and Omer Levy. 2021.
\newblock \href {https://doi.org/10.18653/v1/2021.acl-short.3} {Coreference
  resolution without span representations}.
\newblock In \emph{Proceedings of the 59th Annual Meeting of the Association
  for Computational Linguistics and the 11th International Joint Conference on
  Natural Language Processing (Volume 2: Short Papers)}, pages 14--19, Online.
  Association for Computational Linguistics.

\bibitem[{Lapshinova-Koltunski et~al.(2018)Lapshinova-Koltunski, Hardmeier, and
  Krielke}]{lapshinova-koltunski-etal-2018-parcorfull}
Ekaterina Lapshinova-Koltunski, Christian Hardmeier, and Pauline Krielke. 2018.
\newblock \href {https://aclanthology.org/L18-1065} {{P}ar{C}or{F}ull: a
  parallel corpus annotated with full coreference}.
\newblock In \emph{Proceedings of the Eleventh International Conference on
  Language Resources and Evaluation ({LREC} 2018)}, Miyazaki, Japan. European
  Language Resources Association (ELRA).

\bibitem[{Lee et~al.(2017)Lee, He, Lewis, and Zettlemoyer}]{lee-etal-2017-end}
Kenton Lee, Luheng He, Mike Lewis, and Luke Zettlemoyer. 2017.
\newblock \href {https://doi.org/10.18653/v1/D17-1018} {End-to-end neural
  coreference resolution}.
\newblock In \emph{Proceedings of the 2017 Conference on Empirical Methods in
  Natural Language Processing}, pages 188--197, Copenhagen, Denmark.
  Association for Computational Linguistics.

\bibitem[{Lee et~al.(2018)Lee, He, and Zettlemoyer}]{lee-etal-2018-higher}
Kenton Lee, Luheng He, and Luke Zettlemoyer. 2018.
\newblock \href {https://doi.org/10.18653/v1/N18-2108} {Higher-order
  coreference resolution with coarse-to-fine inference}.
\newblock In \emph{Proceedings of the 2018 Conference of the North {A}merican
  Chapter of the Association for Computational Linguistics: Human Language
  Technologies, Volume 2 (Short Papers)}, pages 687--692, New Orleans,
  Louisiana. Association for Computational Linguistics.

\bibitem[{Luo(2005)}]{luo-2005-coreference}
Xiaoqiang Luo. 2005.
\newblock \href {https://aclanthology.org/H05-1004} {On coreference resolution
  performance metrics}.
\newblock In \emph{Proceedings of Human Language Technology Conference and
  Conference on Empirical Methods in Natural Language Processing}, pages
  25--32, Vancouver, British Columbia, Canada. Association for Computational
  Linguistics.

\bibitem[{Martins(2015)}]{martins-2015-transferring}
Andr{\'e} F.~T. Martins. 2015.
\newblock \href {https://doi.org/10.3115/v1/P15-1138} {Transferring coreference
  resolvers with posterior regularization}.
\newblock In \emph{Proceedings of the 53rd Annual Meeting of the Association
  for Computational Linguistics and the 7th International Joint Conference on
  Natural Language Processing (Volume 1: Long Papers)}, pages 1427--1437,
  Beijing, China. Association for Computational Linguistics.

\bibitem[{Ng et~al.(2019)Ng, Yee, Baevski, Ott, Auli, and
  Edunov}]{ng-etal-2019-facebook}
Nathan Ng, Kyra Yee, Alexei Baevski, Myle Ott, Michael Auli, and Sergey Edunov.
  2019.
\newblock \href {https://doi.org/10.18653/v1/W19-5333} {{F}acebook {FAIR}{'}s
  {WMT}19 news translation task submission}.
\newblock In \emph{Proceedings of the Fourth Conference on Machine Translation
  (Volume 2: Shared Task Papers, Day 1)}, pages 314--319, Florence, Italy.
  Association for Computational Linguistics.

\bibitem[{Nov{\'a}k et~al.(2017)Nov{\'a}k, Nedoluzhko, and
  {\v{Z}}abokrtsk{\'y}}]{novak-etal-2017-projection}
Michal Nov{\'a}k, Anna Nedoluzhko, and Zden{\v{e}}k {\v{Z}}abokrtsk{\'y}. 2017.
\newblock \href {https://doi.org/10.18653/v1/W17-1508} {Projection-based
  coreference resolution using deep syntax}.
\newblock In \emph{Proceedings of the 2nd Workshop on Coreference Resolution
  Beyond {O}nto{N}otes ({CORBON} 2017)}, pages 56--64, Valencia, Spain.
  Association for Computational Linguistics.

\bibitem[{Peters et~al.(2018)Peters, Neumann, Iyyer, Gardner, Clark, Lee, and
  Zettlemoyer}]{peters-etal-2018-deep}
Matthew~E. Peters, Mark Neumann, Mohit Iyyer, Matt Gardner, Christopher Clark,
  Kenton Lee, and Luke Zettlemoyer. 2018.
\newblock \href {https://doi.org/10.18653/v1/N18-1202} {Deep contextualized
  word representations}.
\newblock In \emph{Proceedings of the 2018 Conference of the North {A}merican
  Chapter of the Association for Computational Linguistics: Human Language
  Technologies, Volume 1 (Long Papers)}, pages 2227--2237, New Orleans,
  Louisiana. Association for Computational Linguistics.

\bibitem[{Pradhan et~al.(2014)Pradhan, Luo, Recasens, Hovy, Ng, and
  Strube}]{pradhan-etal-2014-scoring}
Sameer Pradhan, Xiaoqiang Luo, Marta Recasens, Eduard Hovy, Vincent Ng, and
  Michael Strube. 2014.
\newblock \href {https://doi.org/10.3115/v1/P14-2006} {Scoring coreference
  partitions of predicted mentions: A reference implementation}.
\newblock In \emph{Proceedings of the 52nd Annual Meeting of the Association
  for Computational Linguistics (Volume 2: Short Papers)}, pages 30--35,
  Baltimore, Maryland. Association for Computational Linguistics.

\bibitem[{Pradhan et~al.(2012)Pradhan, Moschitti, Xue, Uryupina, and
  Zhang}]{pradhan-etal-2012-conll}
Sameer Pradhan, Alessandro Moschitti, Nianwen Xue, Olga Uryupina, and Yuchen
  Zhang. 2012.
\newblock \href {https://aclanthology.org/W12-4501} {{C}o{NLL}-2012 shared
  task: Modeling multilingual unrestricted coreference in {O}nto{N}otes}.
\newblock In \emph{Joint Conference on {EMNLP} and {C}o{NLL} - Shared Task},
  pages 1--40, Jeju Island, Korea. Association for Computational Linguistics.

\bibitem[{Rahman and Ng(2012)}]{rahman-ng-2012-translation}
Altaf Rahman and Vincent Ng. 2012.
\newblock \href {https://aclanthology.org/N12-1090} {Translation-based
  projection for multilingual coreference resolution}.
\newblock In \emph{Proceedings of the 2012 Conference of the North {A}merican
  Chapter of the Association for Computational Linguistics: Human Language
  Technologies}, pages 720--730, Montr{\'e}al, Canada. Association for
  Computational Linguistics.

\bibitem[{Tiedemann and Thottingal(2020)}]{tiedemann-thottingal-2020-opus}
J{\"o}rg Tiedemann and Santhosh Thottingal. 2020.
\newblock \href {https://aclanthology.org/2020.eamt-1.61} {{OPUS}-{MT} {--}
  building open translation services for the world}.
\newblock In \emph{Proceedings of the 22nd Annual Conference of the European
  Association for Machine Translation}, pages 479--480, Lisboa, Portugal.
  European Association for Machine Translation.

\bibitem[{Toshniwal et~al.(2020)Toshniwal, Wiseman, Ettinger, Livescu, and
  Gimpel}]{toshniwal-etal-2020-learning}
Shubham Toshniwal, Sam Wiseman, Allyson Ettinger, Karen Livescu, and Kevin
  Gimpel. 2020.
\newblock \href {https://doi.org/10.18653/v1/2020.emnlp-main.685} {Learning to
  ignore: Long document coreference with bounded memory neural networks}.
\newblock In \emph{Proceedings of the 2020 Conference on Empirical Methods in
  Natural Language Processing (EMNLP)}, pages 8519--8526, Online. Association
  for Computational Linguistics.

\bibitem[{Vilain et~al.(1995)Vilain, Burger, Aberdeen, Connolly, and
  Hirschman}]{vilain-etal-1995-model}
Marc Vilain, John Burger, John Aberdeen, Dennis Connolly, and Lynette
  Hirschman. 1995.
\newblock \href {https://aclanthology.org/M95-1005} {A model-theoretic
  coreference scoring scheme}.
\newblock In \emph{Sixth Message Understanding Conference ({MUC}-6):
  Proceedings of a Conference Held in {C}olumbia, {M}aryland, November 6-8,
  1995}.

\bibitem[{Wu et~al.(2019)Wu, Fan, Baevski, Dauphin, and Auli}]{wu2019pay}
Felix Wu, Angela Fan, Alexei Baevski, Yann Dauphin, and Michael Auli. 2019.
\newblock \href {https://arxiv.org/abs/1901.10430} {Pay less attention with
  lightweight and dynamic convolutions}.
\newblock In \emph{International Conference on Learning Representations}, New
  Orleans, USA.

\bibitem[{Xia et~al.(2020)Xia, Sedoc, and
  Van~Durme}]{xia-etal-2020-incremental}
Patrick Xia, Jo{\~a}o Sedoc, and Benjamin Van~Durme. 2020.
\newblock \href {https://doi.org/10.18653/v1/2020.emnlp-main.695} {Incremental
  neural coreference resolution in constant memory}.
\newblock In \emph{Proceedings of the 2020 Conference on Empirical Methods in
  Natural Language Processing (EMNLP)}, pages 8617--8624, Online. Association
  for Computational Linguistics.

\bibitem[{Xu and Choi(2020)}]{xu-choi-2020-revealing}
Liyan Xu and Jinho~D. Choi. 2020.
\newblock \href {https://doi.org/10.18653/v1/2020.emnlp-main.686} {Revealing
  the myth of higher-order inference in coreference resolution}.
\newblock In \emph{Proceedings of the 2020 Conference on Empirical Methods in
  Natural Language Processing (EMNLP)}, pages 8527--8533, Online. Association
  for Computational Linguistics.

\end{thebibliography}
\bibliographystyle{acl_natbib}

\appendix 
\section{Experimental Settings}
\label{appdex:setting}

Our experiments are based on the code released by \citet{xu-choi-2020-revealing}.\footnote{\url{https://github.com/lxucs/coref-hoi}} We keep the original settings and do not do hyper-parameter tuning. As \citet{xu-choi-2020-revealing} have shown that higher-order, cluster-level inference does not further boost the performance on coreference resolution given the powerful text encoders, we do not consider higher-order inference in our experiments.  
Even though the mention boundaries are provided in the data, we still let the model learn to detect mentions by itself.  
For evaluation, we follow previous studies and employ the CONLL-2012 official scorer (\citealp{pradhan-etal-2014-scoring}, v8.01)\footnote{\url{https://github.com/conll/reference-coreference-scorers}} 
to compute the F1 scores of three metrics ($MUC$\citep{vilain-etal-1995-model}, $B^{3}$ \citep{bagga1998algorithms}, $CEAF_{e}$\citep{luo-2005-coreference}) and report the average F1 score. 

Regarding the pretrained NMT models, the English-German/French/Russian models are \textit{transformer.wmt19*} and \textit{transformer.wmt14.en-fr} from \url{https://github.com/pytorch/fairseq/blob/main/examples/translation/README.md}, the NMT models for other translation directions are \textit{opus-mt-en-*} or \textit{opus-mt-*en} from \url{https://huggingface.co/Helsinki-NLP}. 

The baseline model is trained on monolingual data while the cross-lingual models are trained on synthetic parallel data. Note that we use the trained monolingual model to initialize the source-side modules of the cross-lingual model. We randomly initialize the parameters of adapters. 
As we train a unified cross-lingual model, we mainly employ cross-lingual pretrained models, the XLM-R base model, as our encoders, but we also explore using two separate monolingual encoders in Section~\ref{ssec:sep-encoders}. 
All the models are trained for 24 epochs with 2 different seeds, and the checkpoint that performs best on the development set is chosen for evaluation. We only report the average scores. Each model is trained on a single NVIDIA V100 GPU with 32GB memory.

\end{document}